\RequirePackage{fix-cm}
\documentclass[smallextended]{svjour3_hal}       
\smartqed  
\usepackage{graphicx}
\usepackage{amssymb,amsmath}
\usepackage{color}
\usepackage{multirow}
\usepackage{booktabs}
\usepackage{color}
\usepackage[utf8]{inputenc}
\usepackage{array}
\usepackage[textsize=tiny]{todonotes}
\usepackage[colorlinks=true,linkcolor=firebrick,citecolor=darkgreen,urlcolor=darkblue]{hyperref}
\usepackage[stable]{footmisc}
\inputencoding{utf8}
\newcolumntype{L}[1]
{>{\raggedright\let\newline\\\arraybackslash\hspace{0pt}}m{#1}}
\newcolumntype{C}[1]
{>{\centering\let\newline\\\arraybackslash\hspace{0pt}}m{#1}}
\newcolumntype{R}[1]
{>{\raggedleft\let\newline\\\arraybackslash\hspace{0pt}}m{#1}}

\definecolor{darkblue}{rgb}{0,0,0.5}
\definecolor{firebrick}{rgb}{0.75,0.125,0.125}
\definecolor{darkgreen}{rgb}{0,0.5,0}
\definecolor{light-gray}{gray}{0.5}

\newcommand\inner[2]{\langle #1, #2 \rangle}

\begin{document}

\title{Similarity encoding for learning with dirty categorical
	variables\thanks{This work was funded by the Wendelin and
DirtyData (ANR-17-CE23-0018) grants.}
}

\author{Patricio Cerda \and
        Gaël Varoquaux \and 
        Balázs Kégl
}

\institute{Inria Saclay \at
          1 Rue Honoré d'Estienne d'Orves, 91120 Palaiseau, France \\
          \email{patricio.cerda@inria.fr}           \and
          LAL, CNRS, France           }

\date{May 30, 2018}

\maketitle

\begin{abstract}
For statistical learning, categorical variables in a table are usually
considered as discrete entities and encoded separately to
feature vectors, e.g., with one-hot encoding. ``Dirty''
non-curated data gives rise to categorical variables with a very high
cardinality but redundancy: several categories reflect the same entity.
In databases, this issue is typically
solved with a deduplication step.
We show that a simple
approach that exposes the redundancy to the learning algorithm brings significant
gains. We study a generalization 
of one-hot encoding, \emph{similarity encoding}, that builds feature
vectors from similarities across categories. We perform a thorough empirical
validation on non-curated tables, a problem seldom studied in machine
learning. Results on seven real-world datasets show that similarity
encoding brings significant gains in prediction
in comparison with known encoding methods for categories or strings, notably
one-hot encoding and bag of character n-grams.
We draw practical
recommendations for encoding dirty categories:
3-gram similarity appears to be a good choice to capture 
morphological resemblance. For very high-cardinality,
dimensionality reduction significantly reduces the computational cost with little
loss in performance: random projections or choosing a subset of prototype categories still outperforms classic encoding approaches.
\end{abstract}

\section{Introduction}
\label{intro}
Many statistical learning algorithms require as input a numerical 
feature matrix. When categorical variables are present in the data, feature
engineering is needed to encode the different categories into a
suitable feature vector\footnote{Some methods, e.g.,
	tree-based, do not require
	vectorial encoding of categories \cite{coppersmith1999partitioning}.}.
One-hot encoding is a simple and widely-used encoding method
\cite{alkharusi2012categorical,berry1998factorial,cohen2013applied,davis2010contrast,pedhazur1973multiple,myers2010research,ogrady1988categorical}.
For example, a categorical variable having as categories
\{\textit{female, male, other}\} can be encoded respectively with 3-dimensional
feature vectors: \{[1, 0, 0], [0, 1, 0], [0, 0, 1]\}.
In the resulting vector space, each category
is orthogonal and equidistant to the others, which agrees with
classical intuitions about nominal categorical variables.

Non-curated categorical data often lead to larger 
cardinality of the categorical variable and give rise to several
problems when using one-hot encoding.
A first challenge is that the dataset may contain
different morphological representations of the
same category.
For instance, for a categorical variable named \textit{company}, it is not
clear if \textit{`Pfizer International
LLC'}, \textit{`Pfizer Limited'}, and \textit{`Pfizer Korea'}
are different names for the same entity, but they are probably related.
Here we build upon the intuition that
these entities should be closer in the feature space than unrelated
categories, e.g., \textit{`Sanofi Inc.'}. 
In \textit{dirty} data, errors such as typos can cause morphological variations
of  the categories\footnote{A detailed taxonomy of
dirty data can be found on Kim \cite{kim2003taxonomy} and a formal description of
data quality problems is proposed by Oliveira \cite{oliveira2005formal}.}.
Without data
cleaning, different string representations of the same category
will lead to completely different encoded vectors.
Another related challenge is that of encoding 
categories that do not appear in the training set.
Finally, with high-cardinality categorical variables, one-hot
encoding can become impracticable due the high-dimensional feature matrix
it creates.

Beyond one-hot encoding, the statistical-learning literature has
considered other categorical encoding methods
\cite{duch2000symbolic,grkabczewski2003transformations,micci2001preprocessing,shyu2005handling,weinberger2009feature},
but, in general, they do not
consider the problem of encoding in the presence of errors, nor how
to encode categories absent from the training set.

From a data-integration standpoint, dirty categories may be seen as a
data cleaning problem, addressed, for instance, with entity resolution.
Indeed, database-cleaning research 
has developed many approaches to curate
categories \cite{pyle1999data,rahm2000data}. Tasks such as
deduplication or record linkage strive to recognize different variants of
the same entity. A classic approach to learning with dirty categories would
be to apply them as a preprocessing step and then proceed with standard
categorical encoding. Yet, for the specific case of supervised learning,
such an approach is suboptimal for two reasons. First, the uncertainty on the
entity merging is not exposed to the statistical model. Second, the
statistical objective function used during learning is not used to guide the entity resolution.
Merging entities is a difficult problem. We build from the assumption that
it may not be necessary to solve it, and that simply exposing similarities
is enough.

In this paper, we study prediction with
high-cardinality categorical variables. We seek a simple
feature-engineering approach to replace the widely used one-hot encoding method.
The problem of dirty categories has not received much attention in the
statistical-learning literature---though it is related to database cleaning
research \cite{krishnan2016activeclean,krishnan2017boostclean}. To ground
it in supervised-learning settings,
we introduce benchmarks on seven real-world datasets
that contain at least one textual categorical variable with a high
cardinality. The goal of this paper is to stress the importance
of adapting encoding schemes to dirty categories by showing that a simple
scheme based on string similarities brings important practical gains.
In \autoref{sec:problem_setting} we describe
the problem of dirty categorical data and its impact on encoding
approaches. In \autoref{sec:related_work}, we describe in detail
common encoding approaches for categorical variables, 
as well as related techniques in database cleaning---record linkage,
deduplication---and in natural language processing (NLP).
Then, we propose in \autoref{sec:similarity_encoding} a softer version
of one-hot encoding, based on string similarity measures.
We call this generalization \emph{similarity encoding}, as it
encodes the morphological resemblance between categories. We also present
dimensionality reduction approaches that decrease the run time of 
the statistical learning task.
Finally, we show in \autoref{sec:empirical_study} the results of a
thorough empirical
study to evaluate encoding methods on dirty categories. On average,
similarity encoding with 3-gram distance is the method that has the best
results in terms of prediction score, outperforming one-hot encoding even
when applying strong dimensionality reduction.

\section{Problem setting: non-standardized categorical variables}
\label{sec:problem_setting}

In a classical statistical data analysis problem, a categorical variable is
typically defined as a variable with values---categories---of either a nominal 
or ordinal nature. For example,
\textit{place of birth} is a nominal categorical variable. Conversely, answers
in the Likert scale to the question: `\textit{Do you agree with this
statement: A child's education is the responsability of parents, not the
school system.}', compose an ordinal categorical variable in which
the level of \textit{agreement} is associated with a numerical value. In
addition, given a prediction problem, variables can be either the target
variable (also known as the dependent or response variable) or an
explanatory variable (a feature or independent variable). In this work, we
focus on the general problem of nominal categorical variables that are part
of the feature set.

In controlled data-collection settings, categorical variables are
standardized: the set of categories is finite and known a
priori---independently from the data---and categories are mutually exclusive.
Typical machine-learning benchmark datasets, as
in UCI Machine Learning Repository, use
standardized categories. For instance, in the Adult
dataset\footnote{\url{https://archive.ics.uci.edu/ml/datasets/adult}.} the 
\textit{occupation} of individuals is described with 14 predefined categories 
in both the training and testing set.

\paragraph{\textbf{A dirty data problem.}} With 
non-standardized categorical variables
the set of possible categories is unknown before the
data collection process. One example of such non-standardized categories
can be found in the Open Payments
dataset\footnote{\url{https://openpaymentsdata.cms.gov/}.}, which
describes financial relationships between healthcare companies
and physicians or teaching hospitals. One possible task is to predict the
value of the binary variable \textit{status} (whether the payment has been
done under a research protocol or not) given the
following variables: \textit{corporation name}, \textit{amount}, and
\textit{dispute} (whether the physician refused the payment in
a second instance). A challenge with this dataset is that some categories
are not standardized. For instance, \autoref{tab:pfizer_freq} shows
all categories of the variable \textit{company name} with the word
\textit{Pfizer} in it for the year 2013.

\begin{table}[tb]
	\caption{Entities containing the word \textit{Pfizer} in the variable
		\textit{company name} of the Open Payments
		dataset (year 2013).}
	\label{tab:pfizer_freq}       
	\centering
	\scriptsize
	\begin{tabular}{lr}
		\hline\noalign{\smallskip}
		\textbf{Company name}                 & \llap{\textbf{Frequency}} 	\\
		\noalign{\smallskip}\hline\noalign{\smallskip}
		Pfizer Inc.                           & 79,073 				\\
		Pfizer Pharmaceuticals LLC            & 486 				\\
		Pfizer International LLC              & 425 				\\
		Pfizer Limited                        & 13 					\\
		Pfizer Corporation Hong Kong Limited  & 4 					\\
		Pfizer Pharmaceuticals Korea Limited  & 3 					\\
		\noalign{\smallskip}\hline
	\end{tabular}
\end{table}

This type of data poses a problem from the point of view of the statistical
analysis because we do not know a priori, without external expert information,
which of these categories refer to the exact same company or whether all
of them have slight differences and hence should be considered as different
entities. Also, we can observe that the frequency of the different categories
varies by several orders of magnitude, which could imply that errors
in the data collection process have been made, unintentionally or not.  

\begin{figure}[tb]
    \begin{minipage}{.3\linewidth}
	\caption{Evolution of the number of categories as a function of the number
		of samples. In six of our seven datasets, a higher number of samples
		implies a higher cardinality of the respective categorical variable.
		The dataset \textit{medical charges} is the only one of this list that
		reaches its highest cardinality (100 categories) at around 1,000
		samples.}
	\label{fig:cardinality_vs_nsamples} 
    \end{minipage}%
    \hfill%
    \begin{minipage}{.68\linewidth}
	\includegraphics[trim={0 0.7cm 0 0cm},clip,width=.95\textwidth]{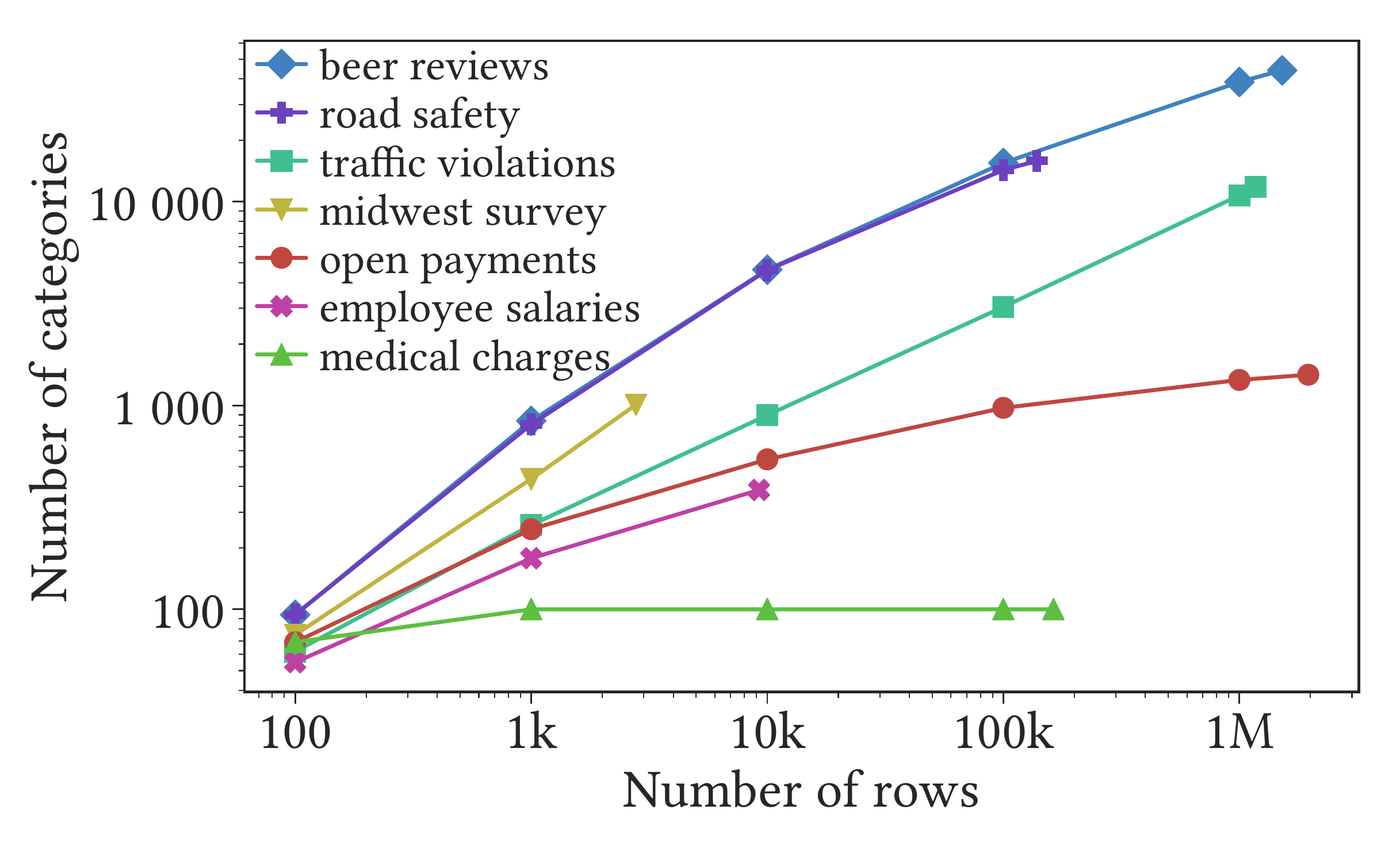}
    \end{minipage}
\end{figure}

Often, the cardinality of a dirty categorical variable
grows with the number of samples in the dataset.
\autoref{fig:cardinality_vs_nsamples} shows the cardinality of the
corresponding categorical variable as a function of the number of samples for
each of the seven datasets that we analyze in this paper.

Dirty categorical data can arise from a variety of mechanisms 
\cite{kim2003taxonomy}:
\begin{itemize}
	\item Typographical errors (e.g., \textit{proffesor} instead of
		\textit{professor})
	\item Extraneous data (e.g., name and title, instead of just the name)
	\item Abbreviations (e.g., \textit{Dr.} for \textit{doctor})
	\item Aliases (e.g., \textit{Ringo Starr} instead \textit{Richard Starkey})
	\item Encoding formats (e.g., ASCII, EBCDIC, etc.)
	\item Uses of special characters (space, colon, dash, parenthesis, etc.)
	\item Concatenated hierarchical data (e.g., state-county-city
	vs. state-city)
\end{itemize}
\paragraph{\textbf{A knowledge-engineering problem.}}
The presence of a large number of categories calls for representing the relationships between them.
In knowledge engineering this is done via an ontology or a taxonomy.
When the taxonomy is unknown, the problem is
challenging. For example, in the \textit{medical charges} dataset,
`cervical spinal fusion' and `spinal fusion except cervical' are different
categories, but both share the fact that they are a spinal fusion,
hence they are not completely independent.

\section{Related work and common practice}
\label{sec:related_work}

Most of the literature on encoding categorical variables relies
on the idea that the set of categories is finite, known a priori, and composed
of mutually exclusive elements \cite{cohen2013applied}. Some studies have
considered encoding high-cardinality categorical variables
\cite{micci2001preprocessing,guo2016entity}, but not the problem of dirty data.
Nevertheless, efforts on this issue have been made in other areas such as
Natural Language Processing and Record Linkage, although they have not been
applied
to encode categorical variables. Below we summarize the main 
relevant approaches.

\paragraph{Notation:} we write sets of elements with capital curly fonts,
as $\mathcal{X}$. Elements of a vector space are written in bold
$\mathbf{x}$, and matrices in capital and bold $\mathbf{X}$. For
a matrix $\mathbf{X}$, we denote by $x^i_j$ the entry on
the $i$-th row and $j$-th column.

\subsection{Formalism: concepts in relational databases and statistical
learning}

We first link our formulations to a
database formalism, which relies on sets.
A table is specified by its \emph{relational scheme} $\mathcal{R}$: the set of
$m$ attribute names $\{A_j, j =1...m\}$, i.e., the column names
\cite{maier1983theory}.
Each attribute name has a domain
$\text{dom}(A_j) = \mathcal{D}_j$. 
A table is defined as a \emph{relation} $r$ on the scheme $\mathcal{R}$: 
a set of
mappings (tuples) $\{t^i: \mathcal{R} \rightarrow \bigcup_{j=1}^{m}
\mathcal{D}_j, \; i=1...n\}$,
where for each \emph{record} (sample)
$t^i \in r$, $t^i(A_j) \in \mathcal{D}_j, \; j = 1...m$.
If $A_j$ is a numerical attribute, then
$\text{dom}(A_j) = \mathcal{D}_j \subseteq \mathbb{R}$.
If $A_j$ is a categorical attribute represented by strings,
then $\mathcal{D}_j \subseteq \mathbb{S}$, where $\mathbb{S}$ is the set of
finite-length strings\footnote{Note that the domain of the categorical
variable depends on the training set.}. As a shorthand, we call
$k_j = \text{card}(\mathcal{D}_j)$ the cardinality of the
variable.

As categorical entities are not numerical, they require
an operation to define a feature matrix $\mathbf{X}$
from the relation $r$. Statistical or machine learning models that need
vector data are applied after a \textbf{categorical variable encoding},
a feature map that consists of
replacing the tuple elements $t^i(A_j), i=1...n$
by feature vectors:
\begin{equation}
\mathbf{x}_j^i \in \mathbb{R}^{p_j}, p_j \geq 1.
\end{equation}
Using the same notation in case of numerical attributes, we can define
$\mathbf{x}_j^i =  t^i(A_j) \in \mathbb{R}^{p_j}, p_j = 1$ and write the
feature matrix $\mathbf{X}$ as:
\begin{equation}
\mathbf{X} = \left[
\begin{array}{*5{c}}
\mathbf{x}_1^1 & \ldots & \mathbf{x}_m^1 \\
\vdots         & \ddots & \vdots \\
\mathbf{x}_1^n & \ldots & \mathbf{x}_m^n
\end{array}\right]
\in \mathbb{R}^{n\times p}, p = \sum_{j = 1}^{m} p_j
\end{equation}
In standard supervised-learning settings, the observations, represented
by the feature matrix $\mathbf{X}$, are associated with a target vector
$\mathbf{y} \in \mathbb{R}^n$ to predict.

We now review classical encoding methods. For simplicity of exposition,
in the rest of the section we will consider only a single categorical 
variable $A$, omitting the column index $j$ from the previous definitions.

\paragraph{\textbf{One-hot encoding.}}
Let $A$ be a categorical variable with cardinality $k \geq 2$ such that
$\text{dom}(A) = \{d_\ell, 1 < \ell \leq k\}$ and $t^i(A) = d^i$.
The one-hot encoding method sets each feature vector as:
\begin{equation}
\mathbf{x}^i = \left[\mathbf{1}_{\{d_1\}}(d^i),\;\; 
		       \mathbf{1}_{\{d_2\}}(d^i),\;\; ...\;, \;\;
		       \mathbf{1}_{\{d_{k}\}}(d^i)
                 \right] \; \in \mathbb{R}^{k}
\label{eq:onehot_encoding}
\end{equation}
where $\mathbf{1}_{\{d_\ell\}}(\cdot)$ is the indicator function over the 
singleton $\{d_\ell\}$.
Several variants of the one-hot encoding have been
proposed\footnote{Variants of one-hot encoding include dummy coding, choosing the 
zero vector for a \textit{reference} category, effects coding, 
contrast coding, and nonsense coding \cite{cohen2013applied}.},
but in a linear regression, all perform equally in terms of $R^2$
score\footnote{The difference between methods is the interpretability
of the values for each variable.} (see Cohen \cite{cohen2013applied} for details).

The one-hot encoding method is intended
to be used when categories are mutually exclusive \cite{cohen2013applied},
which is not necessarily
true of dirty data (e.g., misspelled variables should be
interpreted as overlapping categories).

Another drawback of this method is that it provides no heuristics
to assign a code vector to new categories that appear in the testing set
but have not been encoded on the training set. Given the previous definition,
the zero vector will be assigned to any new category in the testing set, which
creates collisions if more that one new category is introduced.

Finally, high-cardinality categorical variables greatly increase the
dimensionality of the feature matrix, which increases its
computational cost. Dimensionality reduction on the
one-hot encoding vector tackles this problem
(see \autoref{subsec:dimensionality_reduction}), with the risk of loosing
information. 

\paragraph{\textbf{Hash encoding.}}
A solution to reduce the dimensionality of the data is
to use the hashing trick \cite{weinberger2009feature}. Instead of
assigning a different unit vector to each category, as one-hot encoding does,
one could define a hash function to designate a feature
vector on a reduced vector space. This method does not consider the
problem of dirty data either, because it assigns hash values that are
independent of the morphological similarity between categories.

\paragraph{\textbf{Encoding using target statistics.}}
\label{subsec:target-encoding}

The target encoding method \cite{micci2001preprocessing}, is a variation of the
\textit{VDM (value difference metric) continuousification scheme}
\cite{duch2000symbolic}, in which each category is
encoded given the effect it has on the target variable $\mathbf{y}$. The
method considers that categorical variables can contain rare categories.
Hence it represents each category by the probability of $\mathbf{y}$ conditional
on this category. In addition, it takes an empirical Bayes approach to shrink the
estimate. Thus, for a binary classification task:
\begin{equation}
\mathbf{x}^i = \lambda(n^i) \, \mathbb{E}_\ell
\bigl[\mathbf{y}^\ell|d^\ell = d^i \bigr]
+ \bigl(1 - \lambda(n^i) \bigr) \, \mathbb{E}_\ell \bigl[\mathbf{y}^\ell \bigr]
\;\; \in \mathbb{R}
\label{eq:target_encoding}
\end{equation}
where $n^i$ is the frequency of the category $d^i$
and $\lambda(n^i) \in [0, 1]$ is a weight such that its derivative with respect
to $n^i$ is positive, e.g.,
$\lambda(n^i) = (\frac{n^i}{n^i + m}, m > 0$ \cite{micci2001preprocessing}).
Note that the obtained feature vector is in this case one-dimensional.

Another related approach is the MDV continuousification scheme
\cite{grkabczewski2003transformations}, which encodes a category $d^i$ by
its expected value on each target $c_k$,
$\mathbb{E}_\ell \bigl[d^\ell = d^i | \mathbf{y}^\ell = c_k\bigr]$ instead
of $\mathbb{E}_\ell \bigl[\mathbf{y}^\ell|d^\ell = d^i \bigr]$ used in the
VDM. In the case of a classification problem, $c_k$ belongs to the set of
possible classes for the target variable.
However, in a dirty dataset, as with spelling mistakes, some categories can
appear only once, undermining the meaning of their
marginal link to $\mathbf{y}$. 

\paragraph{\textbf{Clustering.}}
To tackle the problem of high dimensionality for high-cardinality categorical
variables, one approach is to
perform a clustering of the categories and generate indicator
variables with respect to the clusters. If  $A$ is a categorical variable
with domain $\mathcal{D}$ and cardinality $k$, we can partition the set
$\mathcal{D}$ into $c \ll k$ clusters
$\mathcal{D}_{1}...\mathcal{D}_{c}$; hence the feature vector
associated to this variable is:
\begin{equation}
\mathbf{x}_j^i = \left[\mathbf{1}_{\mathcal{D}_{1}}(d^i),
\mathbf{1}_{\mathcal{D}_{2}}(d^i), ...,
\mathbf{1}_{\mathcal{D}_{c}}(d^i)\right]
\end{equation}
To build clusters, Micci-Barreca \cite{micci2001preprocessing} proposes grouping categories
with similar target statistics, typically using
hierarchical clustering.

\paragraph{\textbf{Embedding with neural networks.}}
Guo \cite{guo2016entity} proposes an encoding 
method based on neural networks. It is inspired by NLP methods that perform
word embedding based on textual context \cite{mikolov2013efficient}
(see \autoref{subsec:nlp}). In tabular data, the equivalent to this
context is given by the values of the other columns, categorical or not.
The approach is simply a standard neural network, trained to link the
table $\mathcal{R}$ to the target $\mathbf{y}$ with standard
supervised-learning architectures and loss and as inputs the table with
categorical columns one-hot encoded. Yet, Guo \cite{guo2016entity} uses as a 
first hidden layer a bottleneck for each categorical variable.
The corresponding intermediate
representation, learned by the network, gives a vector embedding of the
categories in a reduced dimensionality. This approach is 
interesting as it guides the encoding in a supervised way. Yet, it entails
the computational and architecture-selection
costs of deep learning. Additionally, it is still based on an initial
one-hot encoding which is susceptible to dirty categories.

\paragraph{\textbf{Bag of n-grams.}}
One way to represent morphological variation of strings is 
to build a vector containing the count of all possible n-grams of consecutive characters (or words).
This method is straightforward and naturally creates vectorial
representations where similar strings are close to each other. In this work we
consider n-grams of characters to capture the morphology of short strings.

\subsection{Related approaches in natural language processing}
\label{subsec:nlp}

\paragraph{\textbf{Stemming or lemmatizing.}} Stemming and lemmatizing
are text preprocessing techniques that strive to extract a common root
from different variants of a word \cite{lovins1968development,hull1996stemming}.
For instance,
`standardization', `standards', and `standard' could all be reduced
to `standard'. These techniques are based on a set of rules, crafted to the
specificities of a language. Their drawbacks are that they may not be
suited to a specific domain, such as medical practice, and are costly to
develop. Some recent developments in NLP avoid stemming by working
directly at the character level \cite{bojanowski2016enriching}.

\paragraph{\textbf{Word embeddings.}}
Capturing the idea that some categories are closer than others, such as
`cervical spinal fusion' being closer to `spinal fusion except cervical' than
to `renal failure' in the \emph{medical charges} dataset can be seen as
a problem of learning semantics. Statistical approaches to semantics stem from
low-rank data reductions of word occurrences: the original LSA (latent
semantic analysis) \cite{landauer1998introduction} is a PCA of the
word occurrence matrix in documents; {\tt word2vec} \cite{mikolov2013efficient}
can be seen as a matrix factorization on a matrix of word occurrence in local
windows; and {\tt fastText} \cite{bojanowski2016enriching},
a state-of-the-art approach for
supervised learning on text, is based on a low-rank representation of text.

However, these semantics-capturing embeddings for words cannot
readily be used for categorical columns of a table. Indeed, tabular data
seldom contain enough samples and enough context to train modern
semantic approaches. Pretrained embeddings would not work for
entries drawn from a given specialized domain, such as company names or
medical vocabulary. Business or application-specific tables require
domain-specific semantics.

\subsection{Related approaches in database cleaning}

\paragraph{\textbf{Similarity queries.}}

To cater for different ways information might appear, databases use queries
with inexact matching. Queries using textual similarity
help integration of heterogeneous databases without common domains
\cite{cohen1998integration}.

\paragraph{\textbf{Deduplication, record linkage, or fuzzy matching.}}

In databases, deduplication or record linkage strives to find different
variants that denote the same entity and match them
\cite{elmagarmid2007duplicate}. Classic record
linkage theory
deals with merging multiple tables that have entities in
common. It seeks a combination of similarities across columns and a
threshold to match rows \cite{fellegi1969theory}. If known matching pairs
of entities are available, this problem can be cast as a supervised 
or semi-supervised learning problem \cite{elmagarmid2007duplicate}.
If there are no known matching pairs, the simplest
solution boils down to a clustering approach, often on a similarity
graph, or a related expectation
maximization approach \cite{winkler2002methods}.
Supervising the deduplication task is challenging and often calls for
human intervention. Sarawagi \cite{sarawagi2002interactive} uses active learning to
minimize human effort.
Much of the recent progress in database research strives for faster
algorithms to tackle huge databases \cite{christen2012survey}.

\section{Similarity encoding: robust feature engineering}
\label{sec:similarity_encoding}

\subsection{Working principle of similarity encoding}

One-hot encoding can be interpreted as a
feature vector in which each dimension corresponds to the zero-one
similarity between the category we want to encode and all the known
categories (see \autoref{eq:onehot_encoding}).
Instead of using this particular similarity,
one can extend the encoding to use one of the many string similarities,
e.g., as used for entity resolution. A survey of the most
commonly used text similarity measures can be found in
\cite{cohen2003comparison,gomaa2013survey}.
Most of these similarities are based on a morphological comparison between
two strings. Identical strings will have a similarity equal to 1 and
very different strings will have a similarity closer to 0.
We first describe three of the most commonly used similarity measures:
\paragraph{Levenshtein-ratio.}
It is based on the Levenshtein distance \cite{levenshtein1966binary}
(or edit distance) $d_\text{lev}$ between two strings $s_1$ and $s_2$,
which is calculated as a function of the minimum number
of edit operations that are necessary to transform one string into another.
In this paper we used a Levenshtein distance in which all edit operations have a
weight of 1, except for the \emph{replace} operation,
which has a weight of 2. We obtain a similarity measure using:
\begin{equation}
\text{sim}_{\text{lev-ratio}}(s_1, s_2) = 1 -
\frac{d_\text{lev}(s_1, s_2)}{|s_1|+|s_2|}
\end{equation}
where $|s|$ is the character length of the string $s$. 
\paragraph{Jaro-Winkler.}
\cite{winkler1999state}
This similarity is a variation of the Jaro distance $d_\text{jaro}$
\cite{jaro1989advances}:
\begin{equation}
d_\text{jaro}(s_1, s_2) = \frac{m}{3|s_1|} + \frac{m}{3|s_2|}
+ \frac{m-t}{3m}
\end{equation}
where $m$ is the number of matching characters between $s_1$ and
$s_2$\footnote{Two characters belonging to
	$s_1$ and $s_2$ are considered to be a match if they are identical and the
	difference in their respective positions does not exceed
	$2 \, \text{max}(|s_1|,|s_1|) - 1$.
For m=0, the Jaro distance is set to 0.},
and $t$ is the number of character transpositions between the strings
$s_1$ and $s_2$ without considering the unmatched characters.
The Jaro-Winkler similarity $\text{sim}_\text{j-w}(\cdot, \cdot)$ emphasizes
prefix similarity between the two strings. It is defined as:
\begin{equation}
\text{sim}_\text{j-w}(s_1, s_2) =
1 - \left(d_\text{jaro}(s_1, s_2) + l  p  [1 - d_\text{jaro}(s_1, s_2)]\right)
\end{equation}
where $l$ is the length of the longest common prefix
of $s_1$ and $s_2$, and $p$ is a constant scaling factor.
\paragraph{N-gram similarity.}
It is based on splitting both strings into n-grams and then
calculating the Dice coefficient between them
\cite{angell1983automatic}:
\begin{equation}
\text{sim}_{\text{n-gram}}(s_1, s_2) =
\frac{|\text{n-grams}(s_1) \cap \text{n-grams}(s_2)|}
{|\text{n-grams}(s_1) \cup \text{n-grams}(s_2)|}
\end{equation}
where $\text{n-grams}(s), s \in \mathbb{S},$
is the set of consecutive n-grams for the
string $s$. The notion behind this is that categories sharing a large number of
n-grams are probably very similar.
For instance, $\text{3-grams}(\text{Paris}) =
\{\text{Par}, \text{ari}, \text{ris}\}$ and 
 $\text{3-grams}(\text{Parisian}) =
\{\text{Par}, \text{ari}, \text{ris}, \text{isi}, \text{sia},
 \text{ian}\}$ have three 3-grams in common, and their similarity is 
 $\text{sim}_{\text{3-gram}}(\text{Paris}, \text{Parisian}) = \frac{3}{6}$.

There exist 
more efficient versions of the 3-gram similarity
\cite{kondrak2005n}, but we do not explore them in this work.

\paragraph{\textbf{Similarity encoding.}}
Given a similarity measure, one-hot encoding can be generalized to
account for similarities in categories.
Let $A$ be a categorical variable of cardinality $k$, and let
$\text{sim}: (\mathbb{S} \times \mathbb{S}) \rightarrow [0, 1]$ be an arbitrary
string-based similarity measure so that:
\begin{equation}
\text{sim}(s_1, s_2)  = \text{sim}(s_2, s_1), \quad\forall s_1, s_2 \in \mathbb{S}.
\end{equation}
The similarity encoding we propose replaces the instances of $A$
$d^i, i=1...n$ by a feature vector $\mathbf{x}^i \in
\mathbb{R}^k$ so that:
\begin{equation}
\mathbf{x}^i =  \left[\text{sim}(d^i, d_1), \; \text{sim}(d^i, d_2), \;...,
\;\text{sim}(d^i, d_k)\right] \in \mathbb{R}^k.
\end{equation}

\subsection{Dimensionality reduction: approaches and experiments}
\label{subsec:dimensionality_reduction}

With one-hot or similarity encoding,
high-cardinality categorical variables lead to high-dimensional feature
vectors. This may lead to
computational and statistical challenges.
Dimensionality reduction may be used on the resulting feature matrix.
A natural approach is to use Principal Component Analysis, as it captures
the maximum-variance subspace. Yet, it entails a high computational
cost\footnote{Precisely, the cost of PCA is $\mathcal{O}(n\,p\,\min(n, p))$.}
and is cumbersome to run in a online setting. Hence, we explored
using \textbf{random projections}: based on the Johnson-Lindenstrauss lemma,
these give a reduced representation that accurately approximates distances of
the vector space \cite{rahimi2008random}.

A drawback of such a projection approach is that it requires first
computing the similarity to all categories. Also, it mixes the
contribution of all categories in non trivial ways and hence
may make interpreting the encodings difficult. For this reason, we also explored
prototype based methods: choosing a small number $d$ of categories and
encoding by computing the similarity to these prototypes.
These prototypes should be representative of the full category set in order to have a meaningful reduced space.

One simple approach is to choose the $d \ll k$ \textbf{most frequent 
categories} of the dataset.
Another way of choosing prototype elements in the category set are
clustering methods like \textbf{k-means}, which chooses cluster centers
that minimize a distortion measure. We use as prototype candidates the
closest element to the center of each cluster. Note that we can
apply the clustering on a initial version of the similarity-encoding matrix
computed on a subset of the data. 

Clustering of dirty categories based on a string similarity is strongly
related to deduplication or record-linkage strategies used in database
cleaning. One notable difference with using a cleaning strategy before
statistical learning is that we are not converting the various forms of
the categories to the corresponding cluster centers, but rather
encoding their similarities to these.

\section{Empirical study of similarity encoding}
\label{sec:empirical_study}

To evaluate the performance of our encoding methodology in a prediction task
containing high-cardinality categorical variables, we present an
empirical study on seven real-world datasets. If a
dataset has more than one categorical variable,
only the most relevant one (in terms of predictive
power\footnote{Variables'
predictive power was evaluated with the
feature importances of a Random Forest as implemented in {\tt
scikit-learn} \cite{pedregosa2011scikit}. The feature importance is
calculated as the average (normalized) total reduction of the Gini impurity criterion brought by each feature.})
was encoded with our approach,
while the rest were one-hot encoded.
\begin{table}[tb]
	\caption{Dataset description. The columns \emph{Number of categories},
		\emph{Most frequent category} and \emph{Least frequent category} contain
		information about the particular categorical variable selected for each
		dataset (see \autoref{subsec:datasets} for details)}
	\label{tab:datasets_description}  
	\setlength\tabcolsep{5pt}
	\scriptsize
	\begin{tabular}
		{L{.19\linewidth} R{.11\linewidth} R{.14\linewidth} R{.12\linewidth}
			R{.11\linewidth} L{.15\linewidth}}
		\hline\noalign{\smallskip}
		\textbf{Dataset} &  \textbf{Number of rows} &
		\textbf{Number of categories} & \textbf{Most frequent category} &
		\textbf{Least frequent category}  & \textbf{Prediction type} \\
		\noalign{\smallskip}\hline\noalign{\smallskip}
		medical charges    & 1.6E+05 &  100 & 3023 & 613 &     regression \\
		employee salaries  & 9.2E+03 &  385 &  883 &   1 &     regression \\
		open payments      & 1.0E+05 &  973 & 4016 &   1 &     binary-clf \\
		midwest survey     & 2.8E+03 & 1009 &  487 &   1 & multiclass-clf \\
		traffic violations & 1.0E+05 & 3043 & 7817 &   1 & multiclass-clf \\
		road safety        & 1.0E+04 & 4617 &  589 &   1 &     binary-clf \\
		beer reviews       & 1.0E+04 & 4634 &   25 &   1 & multiclass-clf \\
		\noalign{\smallskip}\hline
	\end{tabular}
\end{table}

\autoref{tab:datasets_description} summarizes the characteristics of 
the datasets and the respective categorical variable
(for more information about the data, see \autoref{subsec:datasets}). The
sample size of the datasets varies from 3,000 to 160,000 and the cardinality of
the selected categorical variable ranges from 100 to more than 4,600 categories.
Most datasets have at least one category that appears only once,
hence when the data is split into a train and test set,
some categories will likely be present only in the testing set.
To measure prediction performance, we use the following metrics: $R^2$ score
for regression, average precision score for binary classification, and
accuracy for multiclass classification. All these scores are upper bounded
by $1$ and higher values mean better predictions.

For the prediction pipeline we used standard data processing and
classification/regression methods implemented in the Python module scikit-learn
\cite{pedregosa2011scikit}. As we focus on evaluating general categorical
encoding methods, all datasets use the same pipeline: no specific parameter
tuning was performed for a particular dataset
(for technical details see \autoref{subsec:prediction_pipeline}).
\begin{figure*}[tb]
	\centering
	\includegraphics[trim={0.5cm 0.5cm 0.5cm 0.7cm},clip,width=.985\textwidth]
	{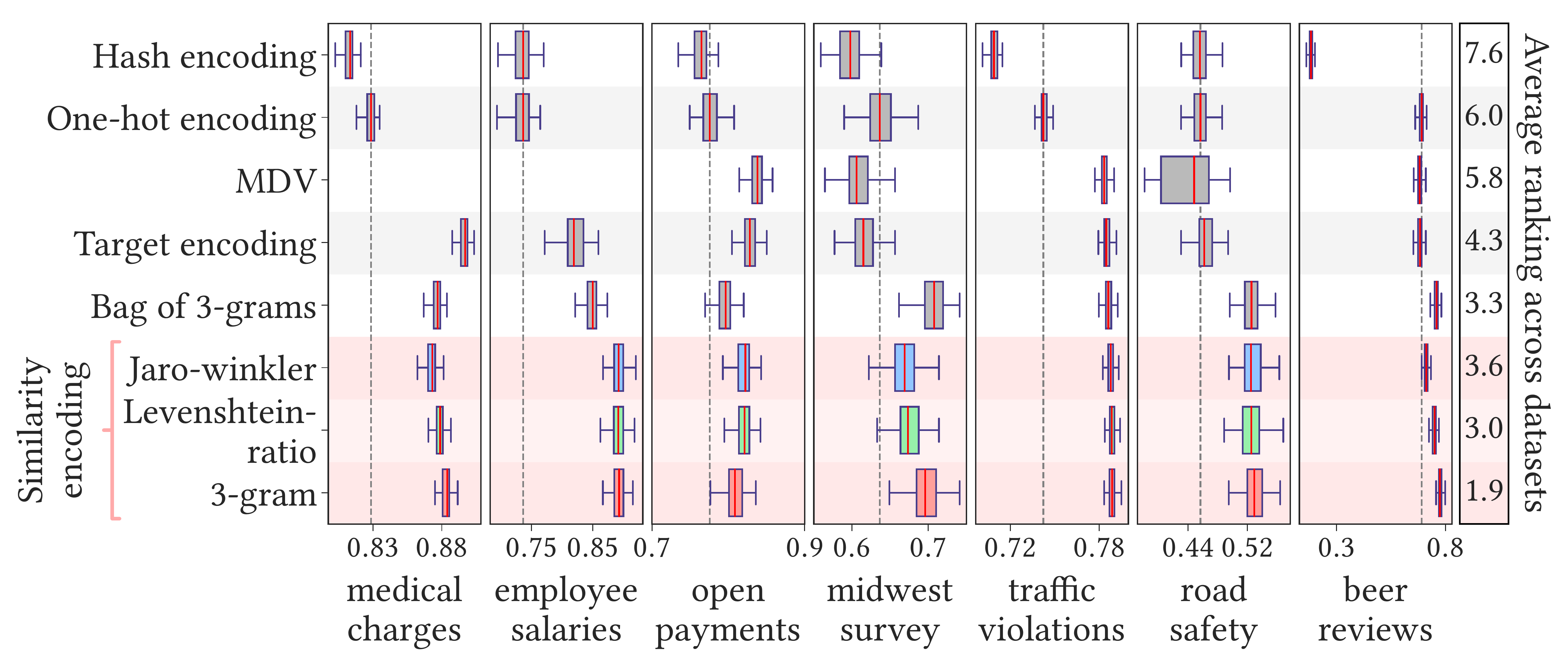}%
	\llap{\rlap{\raisebox{2em}{\sffamily\bfseries
				Gradient}}\hspace*{.98\textwidth}}%
	\llap{\rlap{\raisebox{1em}{\sffamily\bfseries
				boosted trees}}\hspace*{.98\textwidth}}
	\includegraphics[trim={0.5cm 0.5cm 0.5cm 0cm},clip,width=.985\textwidth]
	{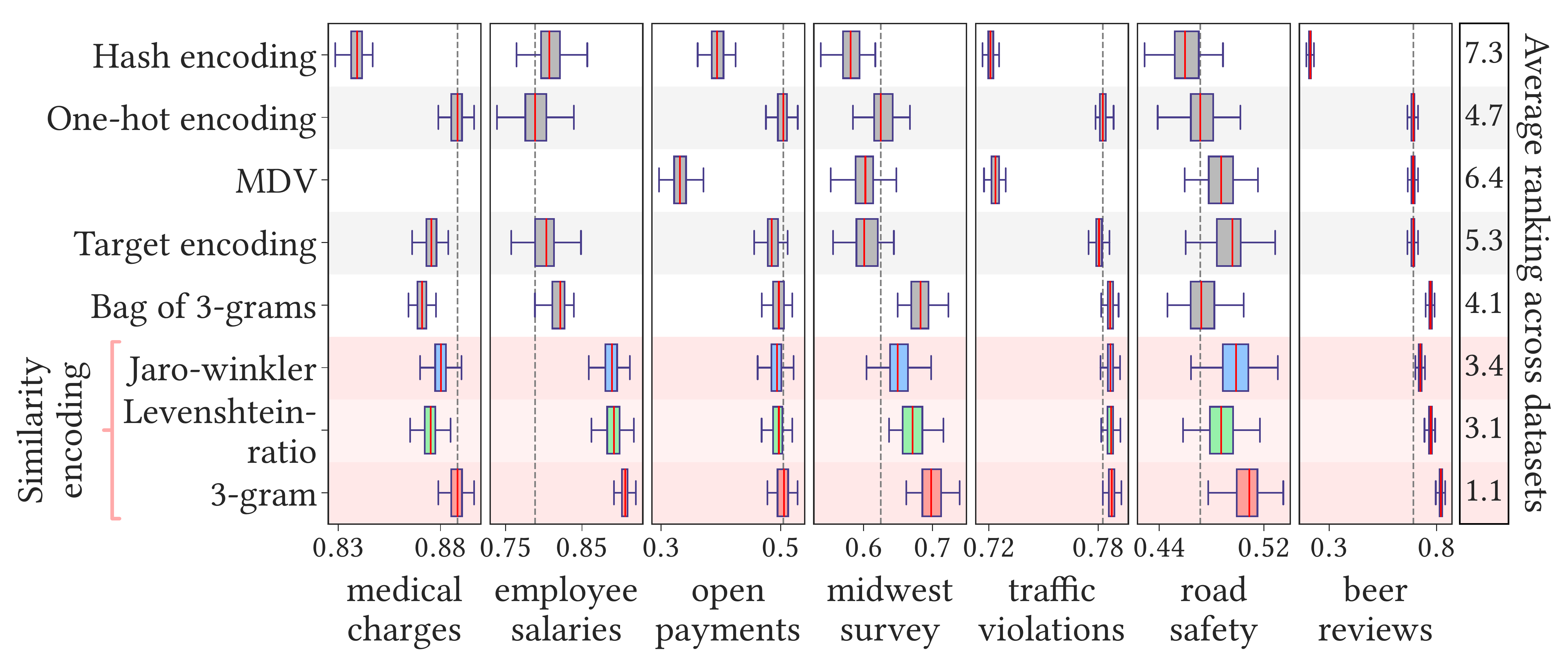}%
	\llap{\rlap{\raisebox{2em}{\sffamily\bfseries
				Ridge}}\hspace*{.98\textwidth}}%
	\llap{\rlap{\raisebox{1em}{\sffamily\bfseries
				regression}}\hspace*{.98\textwidth}}
	
	\caption{\textbf{Performance of different encoding methods.}
		Upper figure: gradient boosting; Lower figure: ridge regression.
		Each box-plot summarizes the prediction scores of 100 random splits
		(with 80\% of the samples for training and 20\% for testing).
		For all datasets, the prediction score is upper bounded by $1$
		(a higher score means a better prediction).
		The right side of the figure indicates the average ranking
		across datasets for each method.
		The vertical dashed line indicates the median value of the one-hot
		encoding method.}
	\label{fig:datasets_distances}
\end{figure*}

First, we benchmarked the similarity encoding with one-hot encoding and
other commonly used methods. Each box-plot in \autoref{fig:datasets_distances}
contains the prediction scores of 100 random
splits of the data (80\% of the samples for training and 20\% for testing)
using gradient boosted trees and ridge regression.
The right side of each plot shows the average ranking of each method
across datasets in terms of the median value of the respective box-plots

In general, similarity encoding methods have the best results in terms of the
average ranking across datasets, with 3-gram being the one that performs
the best for both classifiers (for Ridge, 3-gram similarity is the best
method on every dataset).
On the contrary, the hashing encoder\footnote{We used the MD5 hash function with 256 components.} has the worst performance.
Target and MDV encodings perform well
(in particular with gradient boosting),
considering that the dimension of the feature vector is equal to $1$ for
regression and binary classification, and to the number of classes for
multiclass classification (which goes up to 104 classes for the
\emph{beer reviews}
dataset).
\begin{figure*}[tb]
	\centering
	\includegraphics[trim={0.5cm 0.5cm 0.5cm 0.5cm},clip,width=1\textwidth]
	{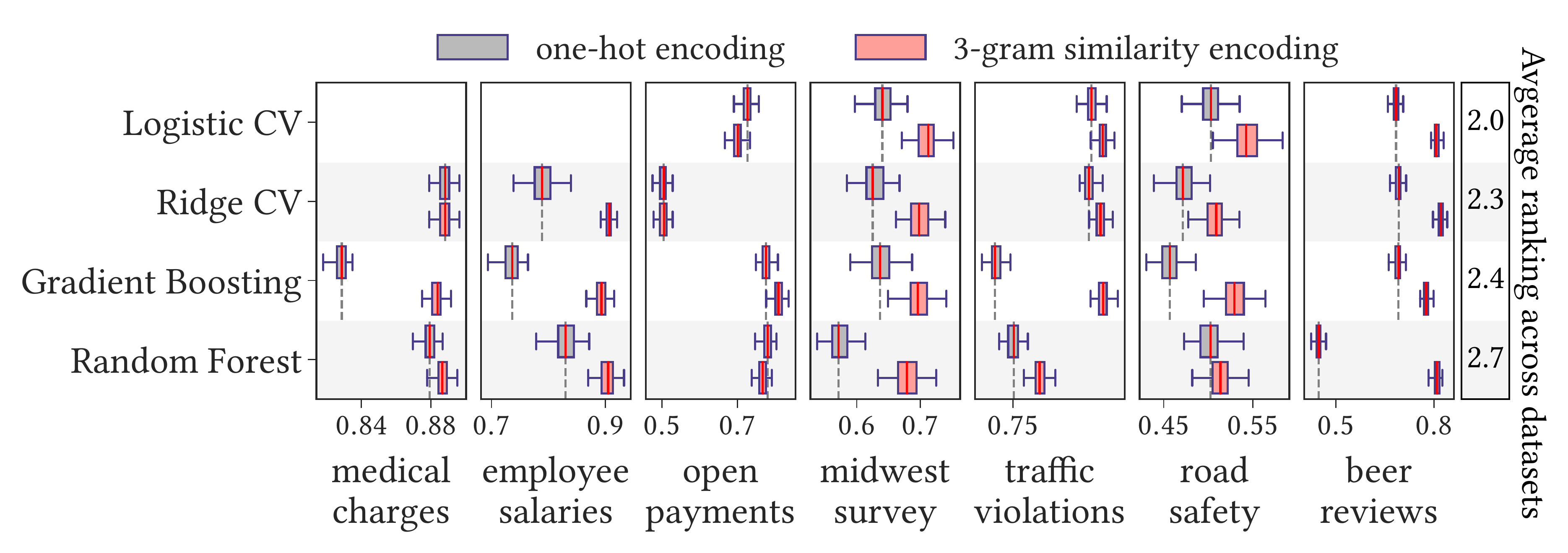}
	\caption{\textbf{Scores with different classifiers} Comparison between
		one-hot and 3-gram similarity encoding. Each box-plot corresponds to 100
		random splits with 20\% of the samples for the testing set.
		The right side of the figure indicates
		the average ranking across datasets for each method in terms of the
		median value of the 3-gram similarity.}
	\label{fig:datasets_3gram_classifiers}       
\end{figure*}

\autoref{fig:datasets_3gram_classifiers} shows
the difference in score between one-hot and similarity encoding for 
different
regressors/classifiers: standard linear methods, ridge
and logistic regression with internal cross-validation of the regularization
parameter, and also the tree-based
methods, random forest and gradient boosting. 
The average ranking is computed with respect to the 3-gram similarity scores.
The \emph{medical charges} and \emph{employee salaries} datasets
do not have scores for the logistic model because their prediction
task is a regression problem. 
\begin{figure*}[tb]
	\centering
	\includegraphics[trim={0.5cm 0cm 0.5cm 0.5cm},clip,width=1\textwidth]
	{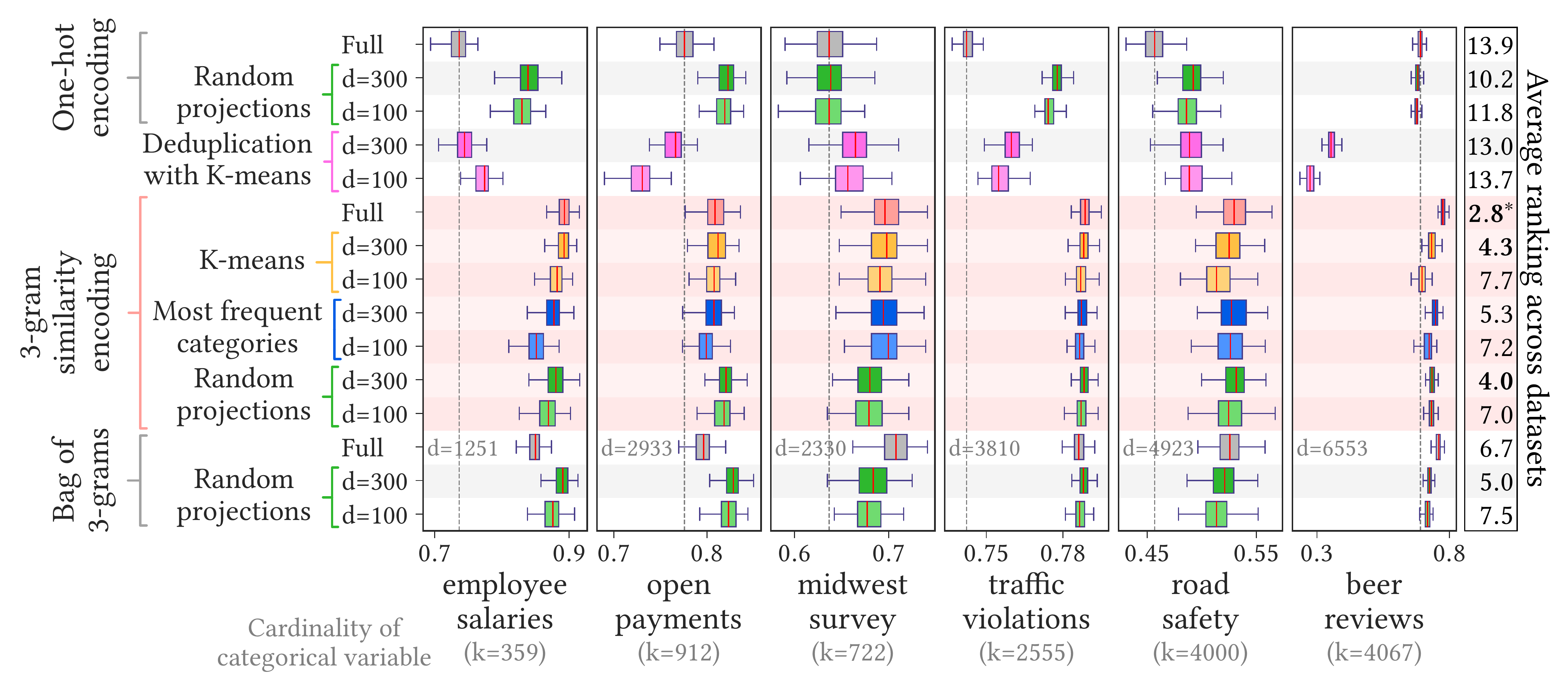}%
	\llap{\rlap{\raisebox{2.9em}{\sffamily\bfseries
			Gradient}}\hspace*{\textwidth}}%
	\llap{\rlap{\raisebox{1.9em}{\sffamily\bfseries
			boosted trees}}\hspace*{\textwidth}}

	\includegraphics[trim={0.5cm 0.5cm 0.5cm 0.5cm},clip,width=1\textwidth]
	{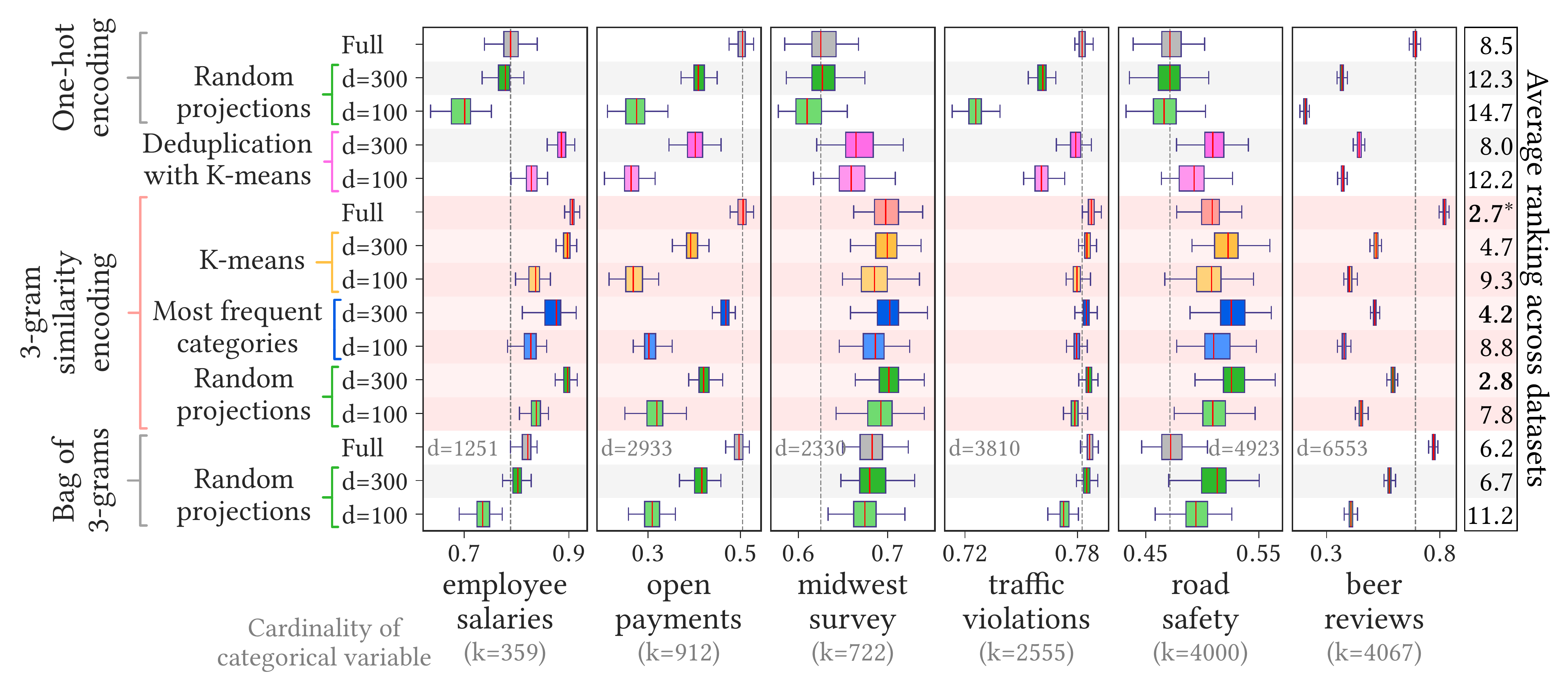}
	\llap{\rlap{\raisebox{2.5em}{\sffamily\bfseries
		Ridge}}\hspace*{1\textwidth}}%
	\llap{\rlap{\raisebox{1.5em}{\sffamily\bfseries
				regression}}\hspace*{1\textwidth}}

	\caption{\textbf{Performance with different dimensionality
			reduction methods}. $Full$ denotes
		the encoding without dimensionality reduction and $d$ the
		dimension of the reduction. Each box-plot corresponds to 100 random
		splits with 80\% of the samples for the training set and 20\% for the
		testing set. The right side of the plot indicates the average
		ranking across datasets for each method ($^*$ denotes the best average ranking).
}
	\label{fig:datasets_dimension-reduction}
\end{figure*}

\autoref{fig:datasets_dimension-reduction} shows
prediction results of different dimensionality reduction methods
applied six of our seven datasets (\emph{medical charges} was
excluded from the figure because of its smaller cardinality in comparison with
the other datasets).
For dimension reduction, we investigated \emph{i)}
random projections, \emph{ii)} encoding with similarities to the 
most frequent categories, \emph{iii)} encoding with similarities to
categories closest to the centers of a k-means clustering,
and \emph{iv)} one-hot encoding after merging categories with a k-means clustering,
which is a simple form of deduplication.
The latter method enables bridging the gap with the deduplication
literature: we can compare merging entities before statistical learning
to expressing their similarity using the same similarity measure.

\section{Discussion}

Encoding categorical textual variables in dirty tables has not been
studied much in the statistical-learning literature. Yet it is a common hurdle
in many application settings. This paper shows that there is room for
improvement upon the standard practice of one-hot encoding by accounting
for similarities across the categories. We studied similarity
encoding, which is a very simple generalization of the one-hot
encoding method\footnote{A Python implementation is available at
\url{https://dirty-cat.github.io/}.}.

An important contribution of this paper is the empirical benchmarks on
dirty tables. We selected seven real-world datasets containing
at least one dirty categorical variable with high-cardinality
(see \autoref{tab:datasets_description}). These datasets are openly
available, and we hope that they will foster more research on dirty
categorical variables. By their diversity, they enable exploring the
trade-offs of encoding approaches and conclude on generally-useful
defaults.

\begin{figure*}[tb]
	\centerline{
		\includegraphics[trim={0 0.5cm 0 0.5cm},clip, width=1\textwidth]{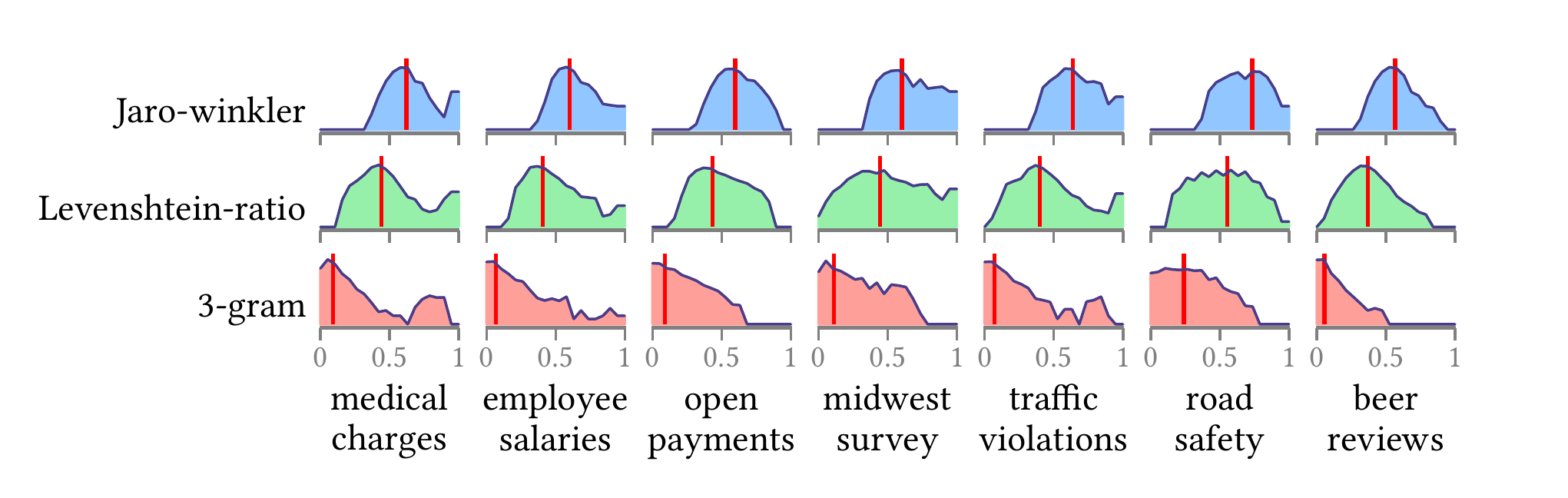}}
	\caption{\textbf{Histogram of pairwise similarity between categories for
			different string similarity metrics.} 10,000 pairs of categories
		were randomly generated for each dataset (y-axis in logarithmic scale).
		The red bar denotes the median value for each distribution. Note that
		\textit{medical charge}, \textit{employee salaries} and
		\textit{traffic violations} present bimodal distributions.}
	\label{fig:histogram_distances}
\end{figure*}

The 3-gram similarity appears to be a good choice,
outperforming similarities typically used for entity
resolution such as Jaro-Winkler and Levenshtein-ratio
(\autoref{fig:datasets_distances}).
A possible reason for the success of 3-gram is visible in the
histogram of the similarities across classes
(\autoref{fig:histogram_distances}).
For all datasets, 3-gram has the smallest median values, and assigns 0
similarity to many pairs of categories. This 
allows better separation of similar and dissimilar categories,
e.g., \emph{`midwest'} and \emph{`mid west'} as opposed to \emph{`southern'}.
3-gram similarity also outperforms the bag of 3-grams. 
Indeed, similarity encoding implicitly defines the following kernel between two observations:
\begin{equation}
\inner{d^i}{d^j}_{\text{sim}} = 
\sum_{l=1}^k \text{sim}(d^i, d_l) \, \text{sim}(d^j, d_l) 
\end{equation}
Hence, it projects on a dictionary of reference n-grams and
gives more importance to the n-grams that best capture 
the similarity between categories.

\autoref{fig:histogram_distances} also reveals that 
three of
the seven datasets (\textit{medical charge}, \textit{employee salaries} and
\textit{traffic violations}) display a 
bimodal distribution in similarities.
On these datasets, similarity encoding brings the largest
gains over one-hot encoding (\autoref{fig:datasets_distances}). In these
situations, similarity encoding is particularly useful as it gives a vector
representation in which a non-negligible number of category pairs are close to each other.

Performance comparisons with different classifiers (linear models and
tree-based models in \autoref{fig:datasets_3gram_classifiers}) suggest
that 3-gram similarity reduces the gap between models by giving a better
vector representation of the categories.
Note that in these experiments linear models slightly outperformed
tree-based models, however we did not tune the hyper parameters of the
tree learners.

While one-hot encoding can be expressed as a sparse matrix, 
a drawback of similarity encoding is
that it creates a dense feature matrix, leading to increased memory
and computational costs. 
Dimensionality reduction of the resulting matrix maintains most of
the benefits of similarity encoding 
(\autoref{fig:datasets_dimension-reduction})
even with a strong reduction ($d$$=$$100$)\footnote{With
	Gradient Boosting, similarity encoding reduced to $d$$=$$30$ still
outperforms one-hot encoding. Indeed, tree models are good at capturing
non-linear decisions in low dimensions.}.
It greatly reduces the computational cost: fitting the models on our benchmark
datasets takes on the order of seconds or minutes on commodity hardware
(see \autoref{tab:prediction_times} in the appendix). Note that on some
datasets, a random projection of one-hot encoded vectors improves
prediction for gradient boosting. We interpret this as a regularization
that captures some semantic links across the categories, as with LSA.
When more than one categorical variable is present, a related approach
would be to use Correspondence Analysis \cite{shyu2005handling}, which
also seeks a low-rank representation as it can be
interpreted as a weighted form of PCA for categorical data. Here we focus
on methods that encode a single categorical variable.

The dimension
reduction approaches that we have studied can be applied in an online
learning setting: they either select a small number prototype categories,
or perform a random projection.
Hence, the approach can be applied on datasets that do not fit in
memory.

Classic encoding methods are hard to apply
in incremental machine-learning settings. Indeed, new samples with new
categories require recomputation of the encoding representation, and hence
retrain the model from scratch.
This is not the case of similarity encoding because new categories are 
naturally encoded without creating collisions. 
We have shown the power of a straightforward strategy 
based on selecting 100 prototypes on subsampled data,
for instance with k-means clustering.
Most importantly, no data cleaning on categorical variables is required to apply
our methodology. Scraped data for commercial or marketing applications are good
candidates to benefit from this approach.

\section{Conclusion}
Similarity encoding, a generalization of the one-hot encoding method,
allows a better representation of categorical variables, especially in the
presence of dirty or high-cardinality categorical data.
Empirical results on seven real-world datasets show that 3-gram similarity
is a good choice to capture morphological resemblance between categories and
to encode new categories that do not appear in the testing set.
It improves prediction of the associated supervised learning task without
any prior data-cleaning step. Similarity encoding also outperforms
representing categories via ``bags of n-grams'' of the associated strings.
Its benefits carry over even with strong
dimensionality reduction based on cheap operations such as random
projections. This methodology can be used in online-learning settings,
and hence can lead to tractable analysis on 
very large datasets without data cleaning.
This paper only scratches the surface of statistical
learning on non-curated tables, a topic that has not been studied much.
We hope that the benchmarks datasets will foster more work on this
subject.

\begin{acknowledgements}
We would like to acknowledge the excellent feedback from the reviewers.
\end{acknowledgements}

\bibliographystyle{spmpsci}      
\bibliography{biblio}   

\section{Appendix}
\label{sec:appendix}

\subsection{Datasets description.}
\label{subsec:datasets}
\paragraph{Medical Charges\footnote{
    \url{https://www.cms.gov/Research-Statistics-Data-and-Systems/Statistics-Trends-and-Reports/Medicare-Provider-Charge-Data/Inpatient.html}}.}
Inpatient discharges for Medicare
beneficiaries: utilization,
payment, and hospital-specific charges for more than 3,000 U.S. hospitals.
Sample size (random subsample): 100,000.
Target variable (regression): `Average total payments'
(what Medicare pays to the provider).
Selected categorical variable:
`Medical procedure' (cardinality: 3023). 
Other explanatory variables:
`State' (categorical),
`Average Covered Charges' (numerical).

\paragraph{Employee Salaries\footnote{
	\url{https://catalog.data.gov/dataset/employee-salaries-2016}}.}
Annual salary information (year 2016) for employees of
Montgomery County, Maryland.
Sample size: 9,200.
Target variable (regression): `Current Annual Salary'.
Selected cat. variable:
`Employee Position Title' (cardinality: 385).
Other explanatory variables:
`Gender' (c),
`Department Name' (c),
`Division' (c),
`Assignment Category' (c),
`Date First Hired' (n).

\paragraph{Open Payments\footnote{
	\url{https://openpaymentsdata.cms.gov}}.}
Payments given by healthcare
manufacturing companies to medical doctors or hospitals. 
Sample size (random subsample): 100,000 (year 2013).
Target variable (binary classification): `Status' (if the payment was made
under a research protocol)
Selected categorical variable: `Company name' (card.: 973).
Other explanatory variables:
`Amount of payments in US dollars' (n),
`Dispute' (whether the physician refused the payment) (c).

\paragraph{Midwest Survey\footnote{
	\url{https://github.com/fivethirtyeight/data/tree/master/region-survey}}.}
Survey
to know if people self-identify as Midwesterners.
Sample size: 2,778.
Target variable (multiclass-clf):
`Location (Census Region)' (10 classes).
Selected categorical variable: `In your own words, what would you call the part
of the country you live in now?' (cardinality: 1,009).
Other explanatory variables:
`Personally identification as a Midwesterner?',
`Gender', `Age', `Household Income', `Education',
`Illinois (IL) in the Midwest?', `IN?', `IA?', `KS?', `MI?',
`MN?', `MO?', `NE?', `ND?',
`OH?', `SD?', `WI?', `AR?',
`CO?', `KY?', `OK?', `PA?',
'WV?', 'MT?', `WY?'.

\paragraph{Traffic Violations\footnote{
	\url{https://catalog.data.gov/dataset/traffic-violations-56dda}}.}
Traffic information from electronic
violations issued in the  Montgomery County of Maryland.
Sample size (random subsample): 100,000.
Target variable (multiclass-clf): `Violation type' (4 classes).
Selected categorical variable: `Description' (card.: 3,043).
Other explanatory variables:
`Belts' (c), `Property Damage' (c), `Fatal' (c),
`Commercial license' (c), `Hazardous materials' (c),
`Commercial vehicle' (c), `Alcohol' (c),
`Work zone' (c), `Year' (n), `Race' (c),
`Gender' (c), `Arrest type' (c).

\paragraph{Road Safety\footnote{
		\url{https://data.gov.uk/dataset/road-accidents-safety-data}}.}
Data
reported to the police about the circumstances of
personal injury road accidents in Great Britain from 1979, and the maker
and model information of vehicles involved in the respective accident.
Sample size (random subsample): 10,000.
Target variable (binary-clf): `Sex of Driver'.
Selected categorical variable: `Model' (card.: 4617)
Other variables: `Make' (c).

\paragraph{Beer Reviews\footnote{
		\url{https://data.world/socialmediadata/beeradvocate}}.}
More than 1.5 million beer reviews. Each review includes ratings in terms of five
``aspects": appearance, aroma, palate, taste, and overall impression.
Sample size (random subsample): 10,000.
Target variable (multiclass-clf): `Beer style' (104 classes).
Selected cat. variable: `Beer name' (card.: 4634)
Other variables (numerical):
`Aroma', `Appearance', `Palate', `Taste'.

\subsection{Technical details on the experiments:
	prediction pipeline\footnote{
	Experiments are available at
	\url{https://github.com/pcerda/ecml-pkdd-2018}}}
\label{subsec:prediction_pipeline}

\paragraph{Sample size.} To reduce computational time on the training step,
we limited the number of samples to 100,000 for large datasets.
For the two datasets with the largest
cardinality of the respective categorical variable (\emph{beer reviews} and
\emph{road safety}), the sample size was set to 10,000.

\paragraph{Data preprocessing.} We removed rows with missing values in the
target variable or in any explanatory variable other than the selected
categorical variable, for which we replaced missing entries by the string `nan'.
The only additional preprocessing for the categorical variable was to transform
all entries to lower case. We standardized every column of the feature matrix to
a unit variance.

\paragraph{Cross-validation.} For every prediction task, we made 100 random
splits of the data, with 20\% of samples for testing at each time. In the
case of binary-class classification, we performed stratified randomization.

\paragraph{Performance metrics.}
Depending on the type of prediction task,
we used different scores to evaluate the performance of the supervised
learning problem:
for regression, we used the $R^2$ score; for binary classification,
the average precision; and for multiclass classification, the accuracy score.

\paragraph{Parametrization of classifiers.}
We used the scikit-learn\footnote{\url{http://scikit-learn.org/}}
implementation of the following methods: LogisticRegressionCV,
RidgeCV (CV denotes internal cross-validation for the regularization parameter),
GradientBoosting and RandomForest. In general, the default parameters
were used, with the following exceptions: \emph{i)} for ensemble methods,
the  number of estimators was set to 100; \emph{ii)} For ridge regression,
we use internal 3-fold cross-validation to set the regularization
parameter; \emph{iii)} when possible, we set
\texttt{class\_weight=`balanced'}. Default parameter settings
can be found at \url{http://scikit-learn.org/}.

\begin{table}[tb]
\scriptsize
\centering
\caption{Average prediction times (in seconds) for the 3-gram similarity encoding
	with k-means for dimensionality reduction.}
\setlength\tabcolsep{5pt} 
\begin{tabular}{lrrrrrrrr}
\hline\noalign{\smallskip}
\multirow{2}{*}{} 
& \multicolumn{4}{c}{\textbf{Gradient boosting}} 
& \multicolumn{4}{c}{\textbf{Ridge CV}} \\
\cmidrule(lr){2-5}\cmidrule(lr){6-9}
\textbf{Dataset} & Full & d=300 & d=100 & d=30 & Full & d=300 & d=100& d=30 \\
\hline\noalign{\smallskip}
Medical charges    &    311 &     - & 156 &   74 &   2.5 &    - &  2.3 & 2.8 \\
Employee salaries  &     69 &    50 &  47 &   37 &   3.9 &  2.8 &  2.6 & 1.6 \\
Open payments      &  1,116 &   393 & 125 &   45 &  61.0 & 12.7 &  2.2 & 0.7 \\
Midwest survey     &    104 &    42 &  14 &  8.6 &   1.9 &  0.4 &  0.1 & 0.1 \\
Traffic violations & 12,165 & 1,686 & 686 &  262 & 116.6 &  7.1 &  2.3 & 1.1 \\
Road safety        &    211 &    30 &  10 &    6 &  78.2 &  1.5 &  0.6 & 0.4 \\
Beer reviews       & 14,214 & 2,260 & 809 &  436 & 302.7 &  2.0 &  0.6 & 0.5 \\
\hline\noalign{\smallskip}         
\end{tabular}
\label{tab:prediction_times}
\end{table}

\end{document}